\documentclass[runningheads,a4paper]{llncs}

\usepackage{amssymb}
\usepackage{graphicx}
\usepackage{color}
\usepackage{algpseudocode}
\usepackage{algorithm}
\usepackage[scriptsize]{subfigure}

\usepackage{url}

\begin{document}

\mainmatter  

\title{Graph-Based Approaches to Clustering Network-Constrained Trajectory Data}

\titlerunning{Clustering Network-Constrained Trajectory Data}

\author{Mohamed K. El Mahrsi\inst{1,2} \and Fabrice Rossi\inst{2}}

\institute{T\'el\'ecom ParisTech, D\'epartement INFRES\\
46, rue Barrault 75634 Paris CEDEX 13, France\\
\email{khalil.mahrsi@telecom-paristech.fr}
\and
\'Equipe SAMM EA 4543, Universit\'e Paris I Panth\'eon-Sorbonne\\
90, rue de Tolbiac 75634 Paris CEDEX 13, France\\
\email{fabrice.rossi@univ-paris1.fr}
}

\authorrunning{Clustering Network-Constrained Trajectory Data}

\maketitle


\begin{abstract}
Clustering trajectory data attracted considerable attention in the last few years. Most of prior work assumed that moving objects can move freely in an euclidean space and did not consider the eventual presence of an underlying road network and its influence on evaluating the similarity between trajectories. In this paper, we present an approach to clustering such network-constrained trajectory data. More precisely we aim at discovering groups of road segments that are often travelled by the same trajectories. To achieve this end, we model the interactions between segments w.r.t. their similarity as a weighted graph to which we apply a community detection algorithm to discover meaningful clusters. We showcase our proposition through experimental results obtained on synthetic datasets.
\end{abstract}

\section{Introduction}

Traffic congestion has become a major problem that affects many human activities on a daily basis, resulting in both serious transportation delays and environmental damages. Monitoring the state of the road network is commonly conducted by using dedicated sensors that register the number of vehicles passing by the section where they are installed. The prohibitive cost of deploying and maintaining such sensors limits their deployment to the highways and the road network's main arteries. Subsequently, the collected data portray a partial and incomplete state of the road network, thus complicating data mining tasks that aim at extracting useful knowledge about flow dynamics and the behavior of drivers moving along the network.

An alternative (or complementary) approach to addressing these shortcomings may consist in analyzing GPS logs collected using location-aware devices (e.g. classic GPS, smartphones, PDAs, etc.). These logs can be acquired through probing vehicles, dedicated data acquisition campaigns (using buses, taxis or an enterprise's fleet of vehicles) or even by means of a crowdsourced approach where different individuals willingly contribute by uploading their different commute logs. Therefore, it is perfectly feasible to collect large amounts of trajectory data that can be stored in dedicated databases (known as Moving Object Databases \cite{Giannotti_2008}). These data offer a better coverage of the road network and can be, later on, explored using data mining and statistical learning techniques.

Clustering is one of such techniques. Prior work on trajectory data clustering focused mainly on the case where moving objects move freely in a euclidean space \cite{Nanni_2006,Benkert_2006,Lee_2007,Jeung_2008a}. By doing so, these approaches did not account for the presence, in the case of car trajectories as well as in other cases, of an underlying network that constrains the movement. The network's constraints, however, do play a paramount role in determining the similarity between the trajectories to be clustered. Moreover, the majority of these approaches relied on the use of density-based clustering which makes them vulnerable to the way the parameters of the clustering algorithm are selected.

In \cite{Mahrsi_2012a}, we presented a framework for clustering network-constrained trajectories using a graph-based approach. This framework was directed towards discovering groups of similar trajectories that moved along the same parts of the road network. The hierarchical, non-parametric algorithm that we used in the clustering step made our framework flexible and suitable for exploring the discovered groups of trajectories at various levels of detail: the user can start with a limited number of high-level, coarse clusters and delve (by means of consecutive zooming) in the refinement of the clusters he deems interesting.

The work presented in this paper builds upon the one undertaken in \cite{Mahrsi_2012a}. We extend our framework to the case of road segments as we try to discover relevant groups of segments that are commonly used and explored together by a considerable number of trajectories. Our contributions can be summarized as follows:
\begin{itemize}
	\item We define a similarity measure that evaluates the resemblance between pairs of road segments based on the trajectories that travelled along both of them;
	\item We use a graph representation to model interactions between the different road segments. The resulting similarity graph is partitioned using modularity-based community detection in order to discover a hierarchy of nested clusters of road segments;
	\item We test our proposition on synthetic datasets and showcase how it can be used, in association with the technique we presented in \cite{Mahrsi_2012a}, for understanding and characterizing the traffic in the road network.
\end{itemize}

The rest of this paper is organized as follows. In Section \ref{sec:datamodel}, we present the network-constrained trajectories data model and we formalize our segment clustering problem. Our segment clustering approach is described in detail in Section \ref{sec:approach}. Section \ref{sec:discussion} discusses the computational complexity of our proposition as well as how the discovered clusters can be interpreted and used, complementarily with the trajectory clusters presented in \cite{Mahrsi_2012a}, in order to discover useful knowledge about the flow dynamics and traffic in the road network. Experimental results are presented in Section \ref{sec:experimentalresults}. Related work is discussed in Section \ref{sec:relatedwork}. Finally, Section \ref{sec:conclusion} concludes the paper.

\section{Data Representation and Problem Statement}
\label{sec:datamodel}

We opt for the symbolic data representation which is the model of choice adopted for representing network-constrained trajectories in most of prior work \cite{Brakatsoulas_2005,Kharrat_2008,Lou_2009,Roh_2010}. In this model, the road network is modeled as a graph, defined as follows.

\begin{definition}[Road Network]
The road network is represented as a directed graph $\mathcal{G} = (\mathcal{V}, \mathcal{S})$. The set of vertices $\mathcal{V}$ represents intersections and terminal points of roads whereas the set of directed edges $\mathcal{S}$ represents the road segments interconnecting them. A directed edge $s = (v_i, v_j)$ indicates that a road segment links the two nodes $v_i$ and $v_j$ and that it can be traveled from $v_i$ in the direction of $v_j$ but not the other way around (unless another edge states otherwise).
\end{definition}

Given this graph representation, moving objects (i.e. vehicles) moving along the road network produce trajectories that can be modeled conformably to the following definition.

\begin{definition}[Constrained Trajectory]
A constrained trajectory $T$ that travels along the road network $\mathcal{G}$ can be modeled as a sequence of visited segments:
\begin{displaymath}
T = \left< id, \left\{s_1, s_2,  ... , s_l\right\} \right>
\end{displaymath}
$id$ being the identifier of the trajectory, $l$ its length (i.e. number of segments) and $\forall 1 \leq i < l, s_i$ and $s_{i+1}$ are connected segments belonging to $\mathcal{S}$.
\end{definition}

In a real-case scenario, trajectories are collected as GPS logs (sequences of latitude and longitude points) on which a map matching technique (e.g. \cite{Brakatsoulas_2005,Lou_2009}) is applied in order to produce the sequence of traveled segments. The map matching step is out of the scope of this paper. Hence, we suppose that the trajectories are already and correctly map matched to the corresponding road segments.

Finally, we formalize the road segment clustering problem that we study in this paper as follows.

\begin{definition}[Road Segment Clustering Problem]
Given a road network represented by a graph $\mathcal{G} = (\mathcal{V}, \mathcal{S})$ and a set of trajectories $\mathcal{T} = \left\{ T_1, T_2, ... , T_n \right\}$ that traveled along it, road segment clustering aims to partition the set of road segments $\mathcal{S}$ into a set of disjoint clusters $\mathcal{C}_\mathcal{S} = \{C_1, C_2, ..., C_K\}$ in such a fashion that:
\begin{itemize}
	\item Segments grouped in the same cluster $C_i$ are visited by a considerable amount of common trajectories (i.e. a trajectory $T$ that visits a segment $s \in C_i$ also visits a fair amount of segments in this same cluster);
	\item Segments belonging to two different clusters $C_i$ and $C_j$ are visited by as few common trajectories as possible (i.e. they are unlikely to be part of a same trajectory).
\end{itemize}
\end{definition}

\section{A Graph-Based Approach to Road Segment Clustering}
\label{sec:approach}

We now present our solution to the road segment clustering problem introduced in the previous section. First, we define a similarity measure between road segments based on the comparison of the common trajectories that visited them (Section \ref{sec:similarity}). Based on this measure, we build a graph depicting the relationships between different road segments (Section \ref{sec:similaritygraph}). The graph is then partitioned using a modularity-based community detection algorithm in order to discover a hierarchy of nested segment clusters (Section \ref{sec:clustering}).

\subsection{Road Segment Similarity}
\label{sec:similarity}

Similarly to the bag-of-words model (where a text is considered as an unordered collection of words), we consider each road segment as a bag-of-trajectories that visited it (i.e. $\forall s \in \mathcal{S}, s \equiv \{T \in \mathcal{T} : s \in T\}$).

In order to compare two road segments $s_i$ and $s_j$, one can simply observe how often they co-appear in trajectories (i.e. calculate $|\{T \in \mathcal{T} : s_i \in T \wedge s_j \in T\}|$). The larger the number of concomitant appearances of both segments is, the more they are considered similar. However, different trajectories do not hold the same discriminative power when it comes to characterizing the similarity between road segments they visit: a lengthy trajectory that travels along a considerable number of road segments is not very informative when judging the similarity between two segments in particular and, vice versa, short trajectories are highly relevant to the formation of the cluster that contains the segments they visit.

We account for this observation by devising a tfidf-like weighting strategy where the contribution of each trajectory is proportional to its length. The weight $\omega_{T, s}$ assigned to trajectory $T$ while inspecting a road segment $s$ is expressed in formula (\ref{eq:weight}):
\begin{equation}
	\omega_{T, s} = \frac{n_{s,T}}{\sum_{T' \in \mathcal{T}} n_{s,T'}} \cdot \log\frac{|\mathcal{S}|}{|s \in \mathcal{S}: s \in T|}
\label{eq:weight}
\end{equation}

The first part in this weight calculates the contribution of $T$ to the segment $s$ by calculating the ratio between the number of appearances $n_{s,T}$ of $s$ in $T$ and the total number of appearances of $s$ in the whole dataset of trajectories $\mathcal{T}$. Since multiple visits of a same road segment are very rare, this part is often equal to $\frac{1}{|\{ T \in \mathcal{T} : s \in T \}|}$. The second part evaluates the importance of the trajectory across the whole set of road segments : the more segments a trajectory visits the less important it becomes and vice versa.

We use a cosine similarity to measure the similarity between two road segments $e_i$ and $e_j$ as expressed in formula (\ref{eq:similarity}):
\begin{equation}
\label{eq:cos}
\mbox{Similarity}(s_i,s_j) = \frac{\sum_{T \in \mathcal{T}} \omega_{T,s_i} \cdot \omega_{T,s_j}}{\sqrt{\sum_{T \in \mathcal{T}} \omega_{T,s_i}^2} \cdot \sqrt{\sum_{T \in \mathcal{T}} \omega_{T,s_j}^2}}
\label{eq:similarity}
\end{equation}

\subsection{Road Segment Similarity Graph}
\label{sec:similaritygraph}

\begin{figure}[htdp]
	\begin{center}
		\includegraphics[width = .7\textwidth]{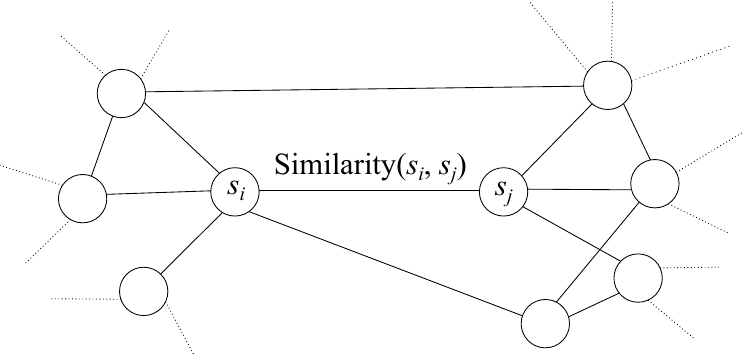}
	\caption{Excerpt from a segment similarity graph. Vertices represent the studied road segments while weighted edges indicate the presence and strength of the similarity between pairs of segments.}
	\label{fig:simgraph}
	\end{center}
\end{figure}

We model the similarity relationships between road segments using an undirected, weighted graph $\mathcal{SG_\mathcal{S} = (\mathcal{S}, \mathcal{E}, \mathcal{W})}$. Each road segment in $\mathcal{S}$ is mapped to a vertice in $\mathcal{SG}_\mathcal{S}$. An edge between a pair of segments $s_i$ and $s_j$ exists if and only if $\mbox{Similarity}(s_i, s_j) > 0$ (i.e. if there is at least one common trajectory that crossed both segments). In which case the similarity is assigned as a weight to that edge. This concept of similarity graph is depicted in Fig. \ref{fig:simgraph}.

The main advantage of using this graph representation, besides being natural and easy to understand, is that it does not invent an "artificial" similarity between totally incompatible road segments. On the contrary, it emphasizes on the fact that road segments that do not share common trajectories are independent and should, therefore, not be "immediately" grouped in the same cluster since there is no similarity edge linking them.

\subsection{Clustering the Similarity Graph}
\label{sec:clustering}

Road networks are complex and contain a considerable amount of segments, resulting, therefore, in a large similarity graph. Moreover, since one common trajectory is sufficient for a similarity edge to exist between a pair of segments, the vertices of the similarity graph tend to have high degrees (although, from our observations, this degree distribution does not follow a proper power law). Modularity-based community-detection algorithms are a popular and widely adopted choice to clustering such graphs \cite{Fortunato_2010}.

Given a graph $\mathcal{G} = (\mathcal{V}, \mathcal{E}, \mathcal{W})$, with vertices $\mathcal{V} = \{v_1, v_2, ... , v_n\}$, weighted edges $\mathcal{E}$ such as $\omega_{ij} \geq 0$ and $\omega_{ij} = \omega_{ji}$, and given a partition of the vertices into $K$ clusters (or communities) $C_1, ... , C_K$, the modularity of the partition is expressed according to formula (\ref{eq:modularity}):

\begin{equation}
\mathcal{Q} = \frac{1}{2m}\sum_{k = 1}^K \sum_{i, j \in C_k} \left( \omega_{ij} - \frac{d_i d_j}{2m} \right)
\label{eq:modularity}
\end{equation}
$d_i = \sum_{j \neq i} \omega_{ij}$ and $m = \frac{1}{2}\sum_i d_i$. The modularity measures the quality of the clustering by inspecting the arrangement of the edges within the communities of vertices. A high modularity is an indicator that the edges within the communities outnumber (or have higher weights than) those in a similar randomly generated graph (i.e. one that does not present a community structure). Communities discovered using modularity optimization have a structure that is similar to the structure of cliques. In our context of segment clustering, this means that segments grouped together are heavily connected (which is the intended result) and are travelled by a considerable number of shared trajectories.

To cluster the segment similarity graph, we use the implementation of hierarchical modularity-based clustering described in \cite{Rossi_2011}. The pseudo-code is given in Algorithm \ref{alg:clustering}. First, the algorithm retrieves a partition of the vertices with optimal modularity (line 1): the \textit{Partition} procedure start by considering the trivial partition where each vertex is in its own community and merges communities in a greedy fashion (i.e. each time, it merges the two communities that produce the maximum increase of modularity). The merging operation stops when no possible merge can be done without a degradation of the modularity. In which case the \textit{Partition} procedure proceeds to a refinement step where members of different communities are interchanged in an attempt to further improve the modularity of the partition.

\begin{algorithm}[!ht]
\begin{algorithmic}[1]
\renewcommand{\algorithmicrequire}{\textbf{Input:}}
\renewcommand{\algorithmicensure}{\textbf{Output:}}
\Require an undirected, weighted graph $\mathcal{G} = (\mathcal{V}, \mathcal{E}, \mathcal{W})$
\Ensure hierarchy of nested clusters of vertices

\State $C_1^{(1)}, ... C_K^{(1)} \leftarrow$ \textit{Partition}($\mathcal{G}$) \Comment{initial partition}
\State $K_T \leftarrow K$ \Comment{clusters counter}
\State $l \leftarrow 1$ \Comment{hierarchy level}

\Repeat
\State $l \leftarrow l + 1$
\ForAll{cluster $C \in C_1^{(l-1)}, ..., C_K^{(l-1)} $}
\State extract the sub-graph $\mathcal{G}_C$ of vertices belonging to $C$
\State $C_1^{C}, ... C_k^{C} \leftarrow$ \textit{Partition}($\mathcal{G}_C$)
\If{\textit{TestSig}($C_1^{C}, ... C_k^{C} $)}
\State $C_{K_T+1}^{(l)}, ... , C_{K_T+k}^{(l)} \leftarrow C_1^{C}, ... C_k^{C}$
\State $K_T \leftarrow K_T + k$
\EndIf
\EndFor
\Until{no significant subdivision of level $l$ can be found}
\end{algorithmic}
\caption{Hierarchical modularity-based clustering.}
\label{alg:clustering}
\end{algorithm}

Once the initial partition is retrieved, the algorithm proceeds iteratively to construct the hierarchy of communities (lines 4 through 14). For each community at a given level, the sub-graph containing only the vertices of the community and the edges connecting them is isolated (line 7). This subgraph is partitioned separately (by invoking \textit{Partition} as shown in line 8). The \textit{TestSig} evaluates the significance of the found partition (by comparing its modularity to the modularity of partitions obtained on similar randomly generated graphs). If the partition is significant indeed, its communities are considered for partitioning in the next iteration (lines 9-12), otherwise it's rejected and the original community is retained. The iterations stop when none of the communities at level $l$ yield a significant partition (line 14).

Modularity-based graph clustering approaches are very popular and achieve good results in practice \cite{Fortunato_2010}. Nevertheless, we do not exclude the use of other graph clustering alternatives (e.g. spectral clustering \cite{Luxburg_2007}) if such techniques can yield better results.

\section{Discussion}
\label{sec:discussion}

First, we focus on how the produced clusters can be explored and analyzed in order to deduce useful knowledge about the flow dynamics and the drivers' behavior in the road network (Section \ref{sec:exploration}). Then, we address the algorithmic complexity of our approach (Section \ref{sec:complexity}).

\subsection{Cluster Exploration}
\label{sec:exploration}

\begin{figure}[!ht]
\centering
	\subfigure[Cluster 1 (14 trajectories)]{
		\includegraphics[scale=0.21]{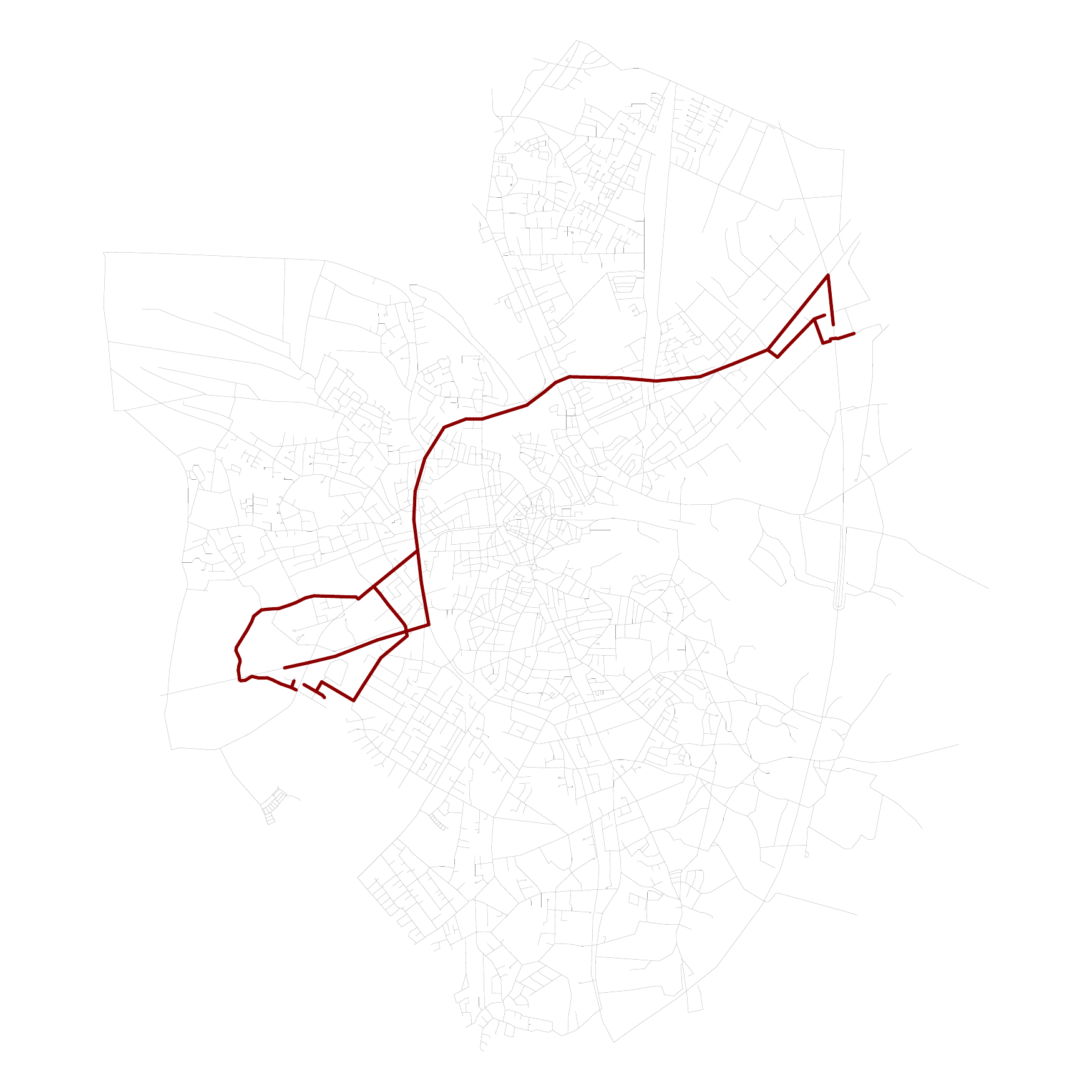}
	}
	\subfigure[Cluster 2 (19 trajectories)]{
		\includegraphics[scale=0.21]{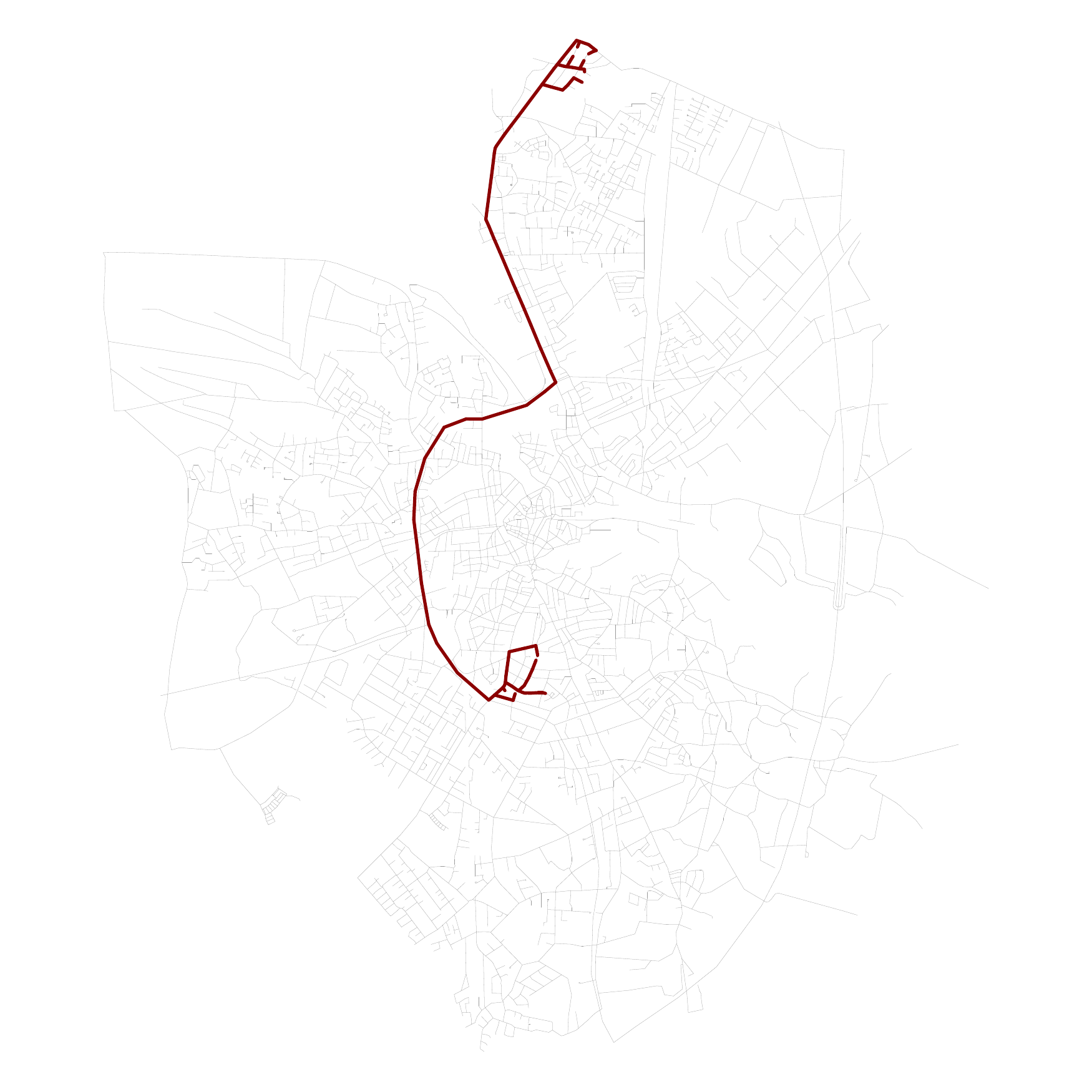}
	}
	\subfigure[Cluster 3 (20 trajectories)]{
		\includegraphics[scale=0.21]{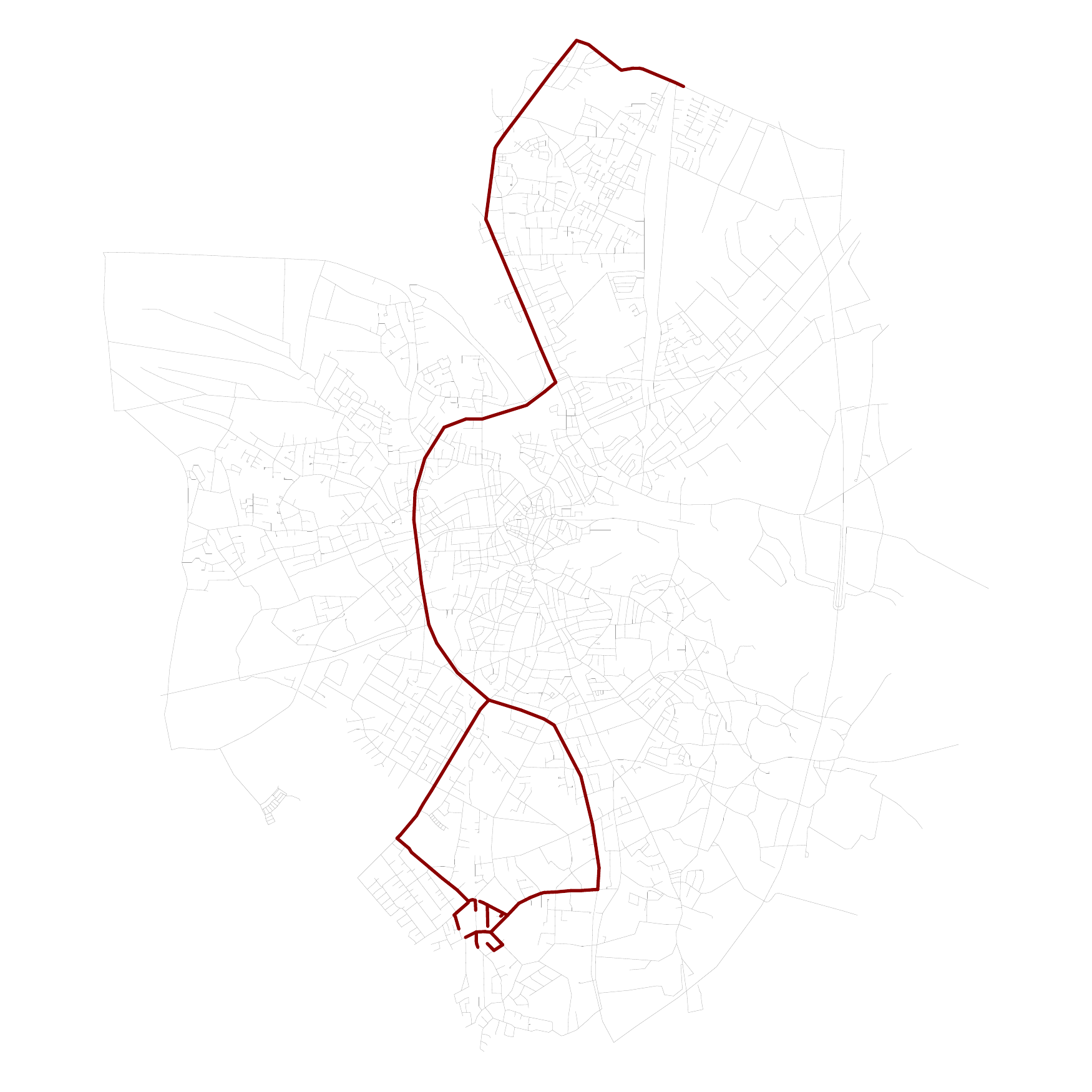}
	}
	\subfigure[Cluster 4 (20 trajectories)]{
		\includegraphics[scale=0.21]{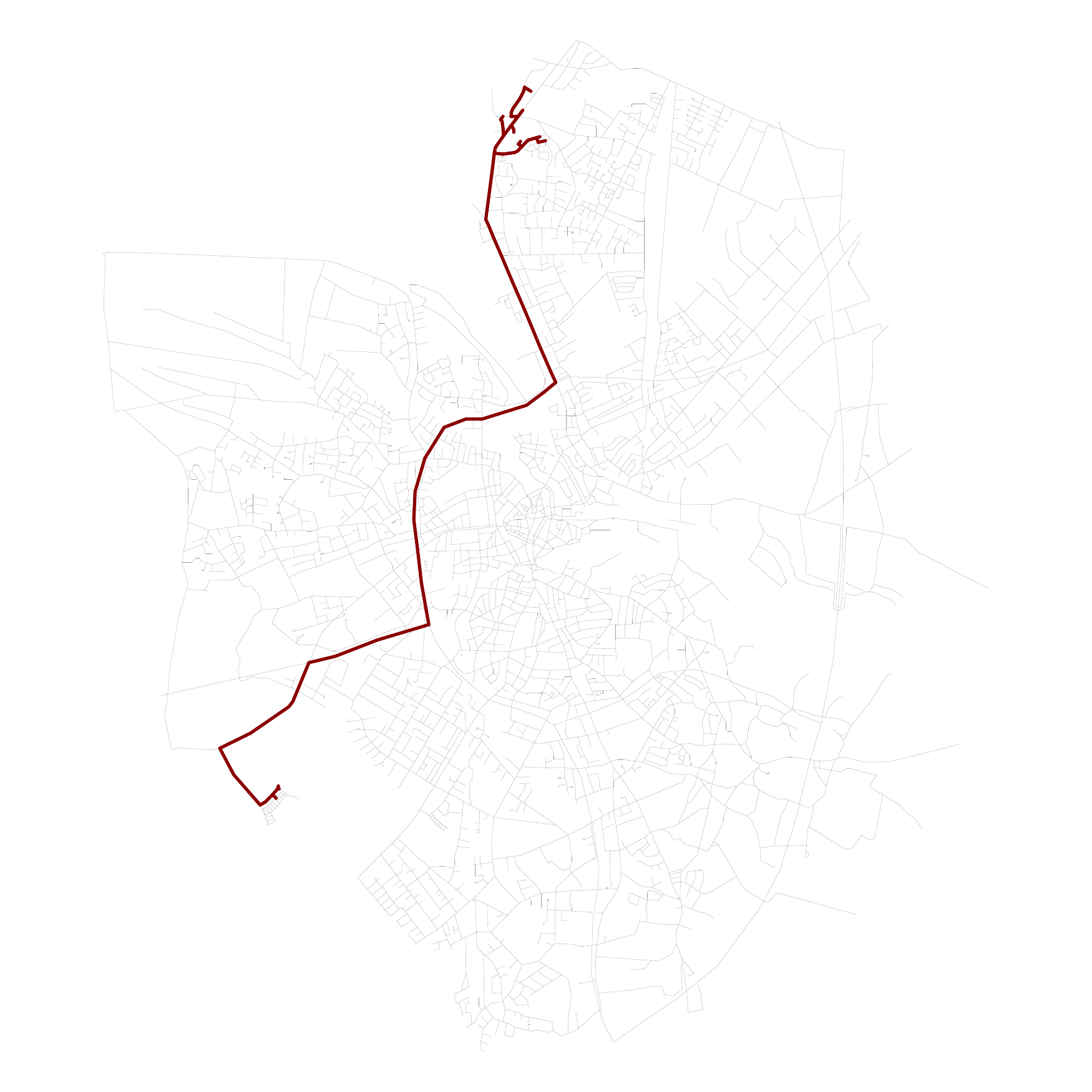}
	}
	\subfigure[Cluster 5 (12 trajectories)]{
		\includegraphics[scale=0.21]{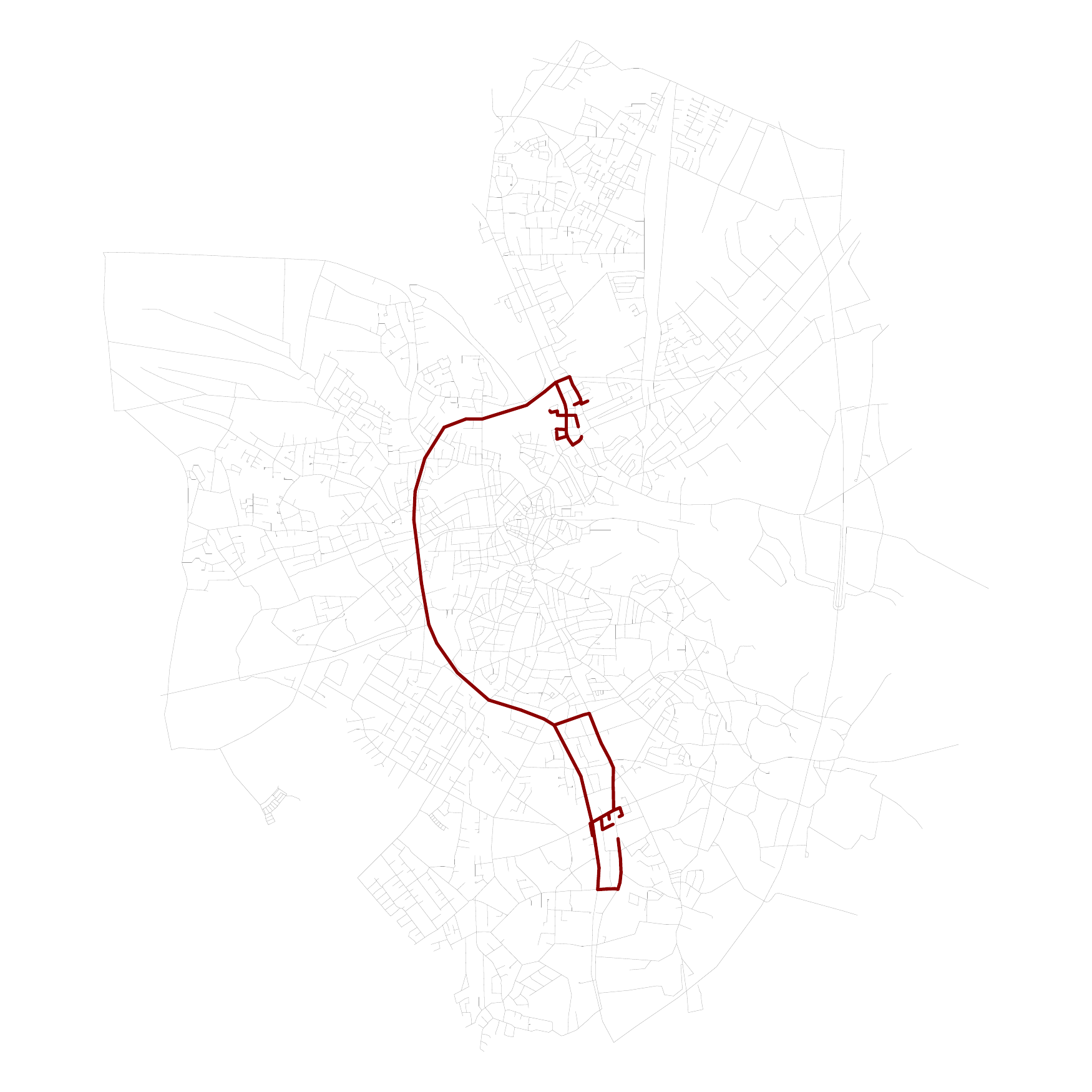}
	}
\caption{Ground-truth clusters in the dataset.}
\label{fig:clusters}
\end{figure}

We illustrate how the trajectory clusters \cite{Mahrsi_2012a} and segment clusters can be explored and used conjointly.
For illustration purposes, we use a synthetic dataset containing 85 trajectories that moved along the Oldenburg road network (cf. Section \ref{sec:experimentalresults} for more details about this network) and visited a total of 485 distinct road segments. We manually partitioned the trajectories into five clusters (depicted in Figure \ref{fig:clusters}) that we consider hereafter as the ground-truth clusters.

\begin{figure}[!ht]
\centering
	\subfigure[Parent cluster (39 trajectories)]{
		\includegraphics[scale=0.22]{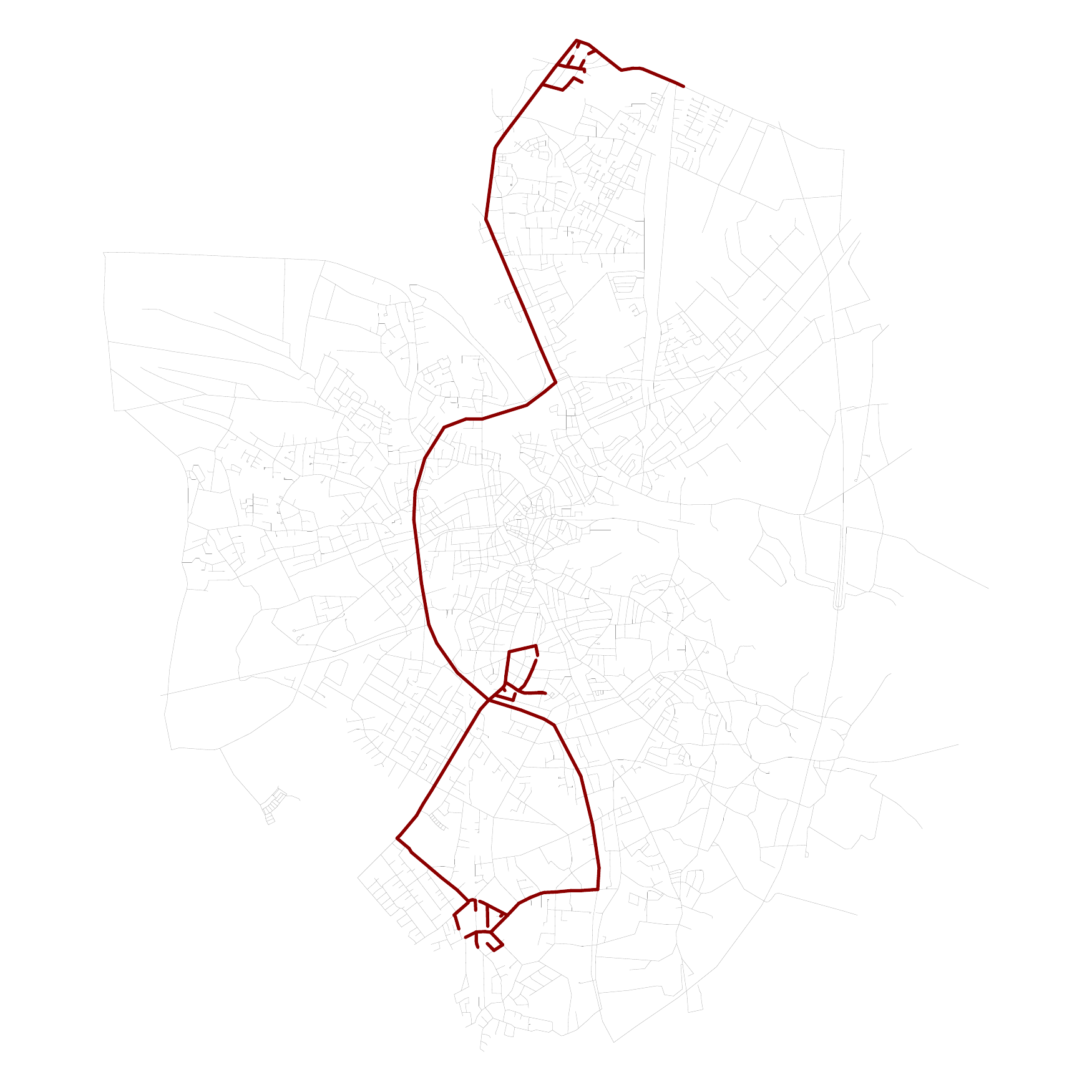}
	}
	\subfigure[Subcluster 1 (12 trajectories)]{
		\includegraphics[scale=0.22]{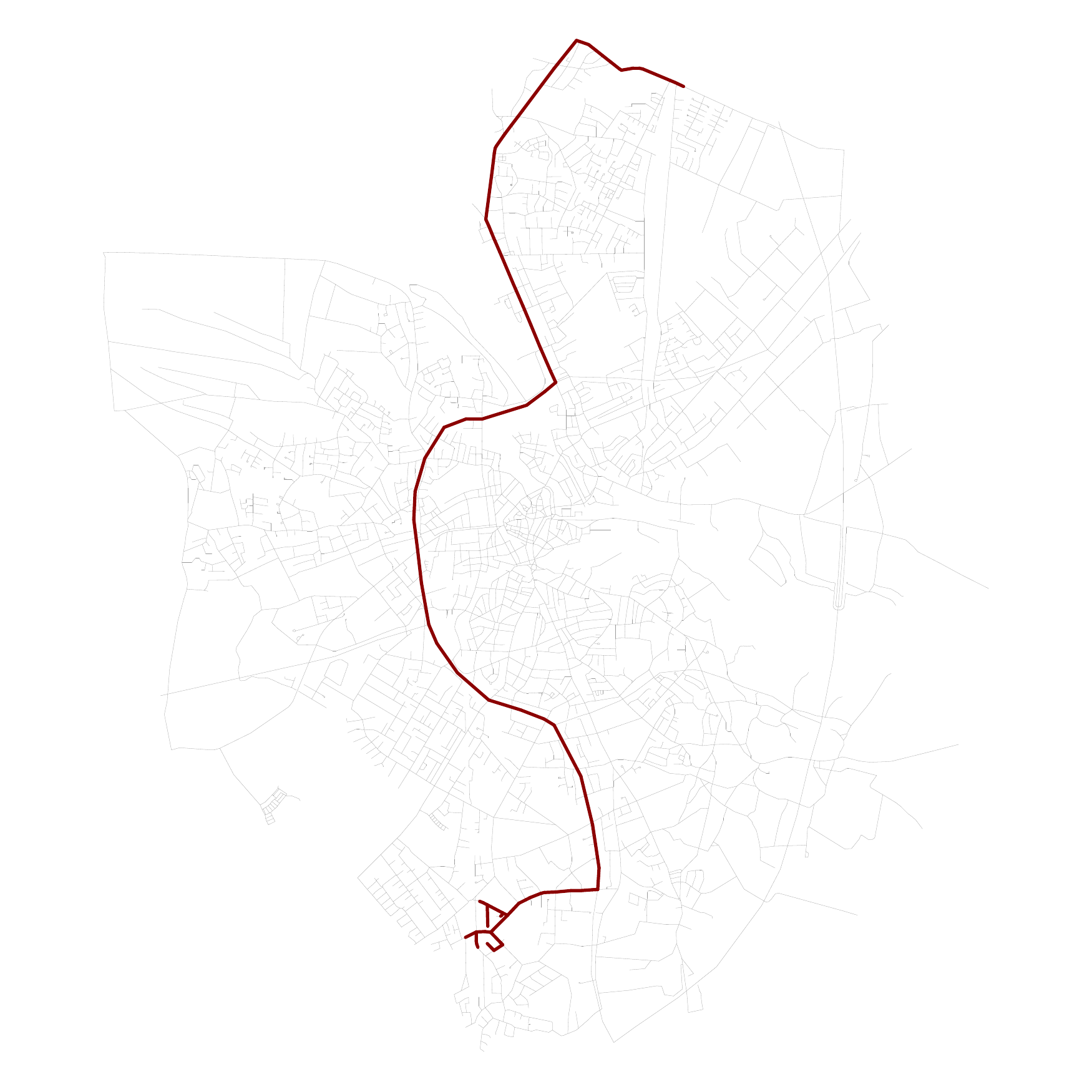}
	}
	
	\subfigure[Subcluster 2 (19 trajectories)]{
		\includegraphics[scale=0.22]{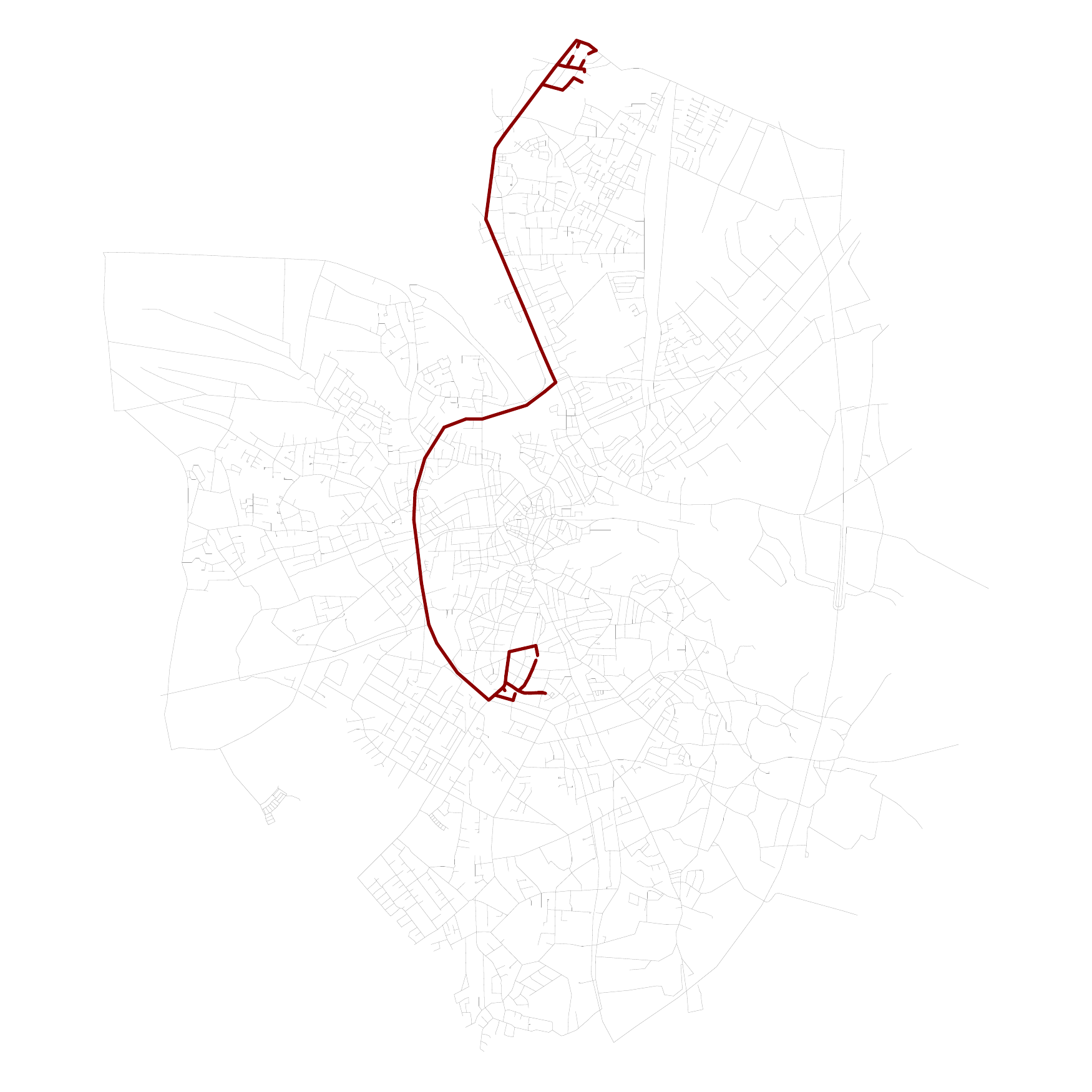}
	}
	\subfigure[Subcluster 3 (8 trajectories)]{
		\includegraphics[scale=0.22]{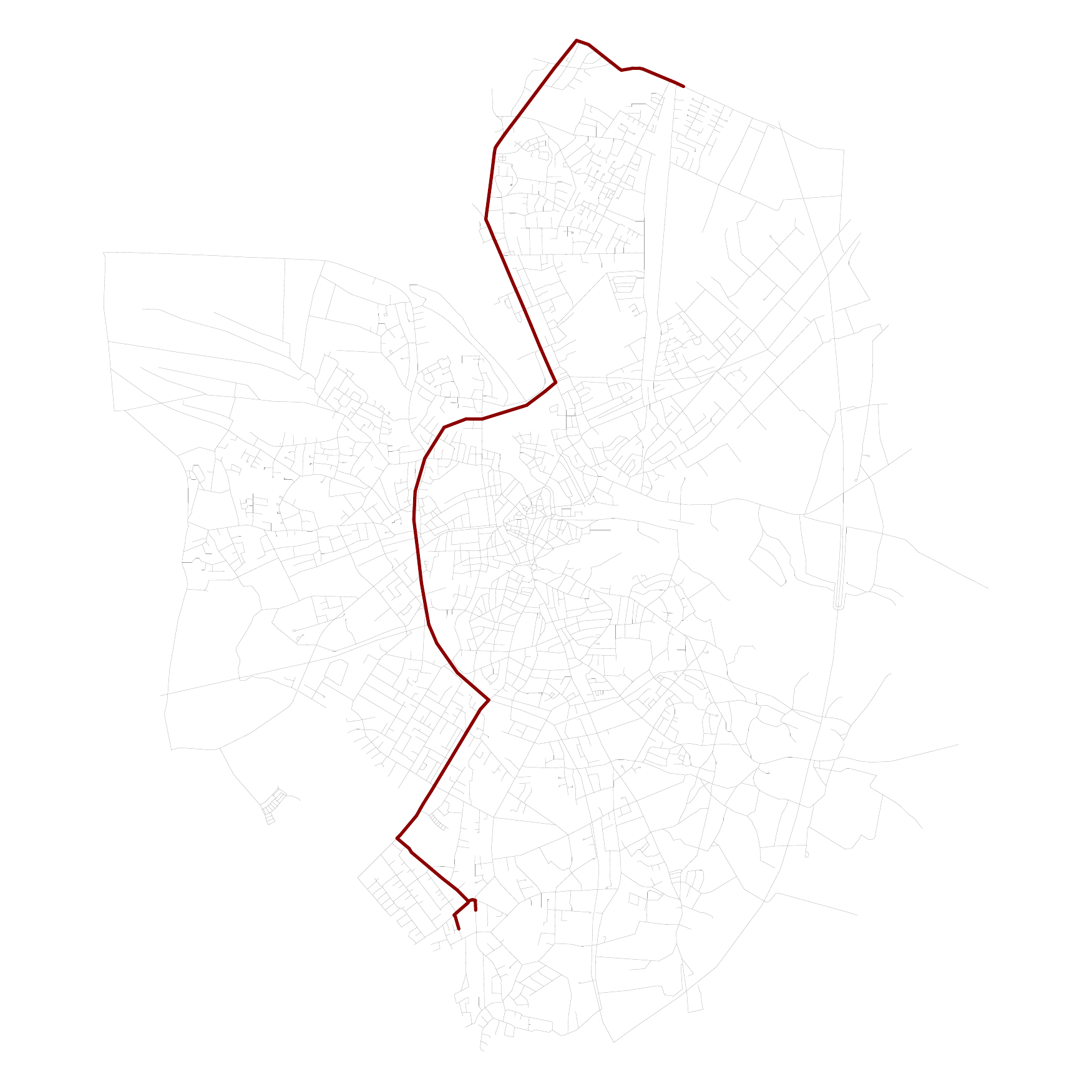}
	}
\caption{A coarse cluster containing 39 trajectories (a) and its more detailed subclusters (b-d).}
\label{fig:zooms}
\end{figure}

Applying the trajectory clustering \cite{Mahrsi_2012a} results in a hierarchy of clusters where the optimal level w.r.t. modularity (i.e. the very first level) contains only three trajectory clusters: the ground-truth clusters 2 and 3 are considered as part of a same cluster (the same occurs with clusters 4 and 5). Nevertheless, all the ground-truth clusters are retrieved correctly (some of them are even refined) in the following levels. The cluster hierarchy is especially suitable for exploring large datasets where a flat clustering can still produce a high number of clusters: the analyst can start with the few, coarse clusters contained in the first hierarchical levels in order to gain a quick grasp of the general tendencies and movement patterns in the road network. He, then, can choose clusters of interest that he can explore, by means of successive zooms, in higher detail. This idea is depicted in Figure \ref{fig:zooms} which shows a coarse trajectory cluster and its three, more refined subclusters.

\begin{figure}[!ht]
	\begin{center}
		\includegraphics[width = \textwidth]{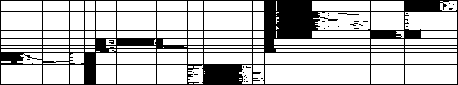}
	\caption{Crossed matrix of the trajectory clusters (rows) and road segment clusters (columns). Each cell gives an idea about the interaction between the corresponding trajectory and segment clusters: the more black the cell contains the more trajectories in the trajectory cluster cross segments belonging to the segment cluster.}
	\label{fig:crossedmatrix}
	\end{center}
\end{figure}

Segment clusters are not as easy to grasp and understand as trajectory clusters. Even though it is feasible to try  
and explore these clusters as stand-alone clusters, we recommend involving the trajectory clusters in the process. Cross-comparing both types of clusters can reveal interesting information about flow dynamics and yield a better interpretation of the clusters. For example, a segment cluster can be interpreted based solely on the trajectory groups that interacted with it, thus revealing potential hubs, etc. Fig. \ref{fig:crossedmatrix} shows the crossed matrix of the second level trajectory clusters (reported on the rows) and the second level road segment clusters (on the columns) and gives an idea about the sizes of the clusters and how clusters of one type interact with those of the other type.

\begin{figure}[!ht]
\centering
	\subfigure[Hub segment cluster]{
		\includegraphics[scale=0.21]{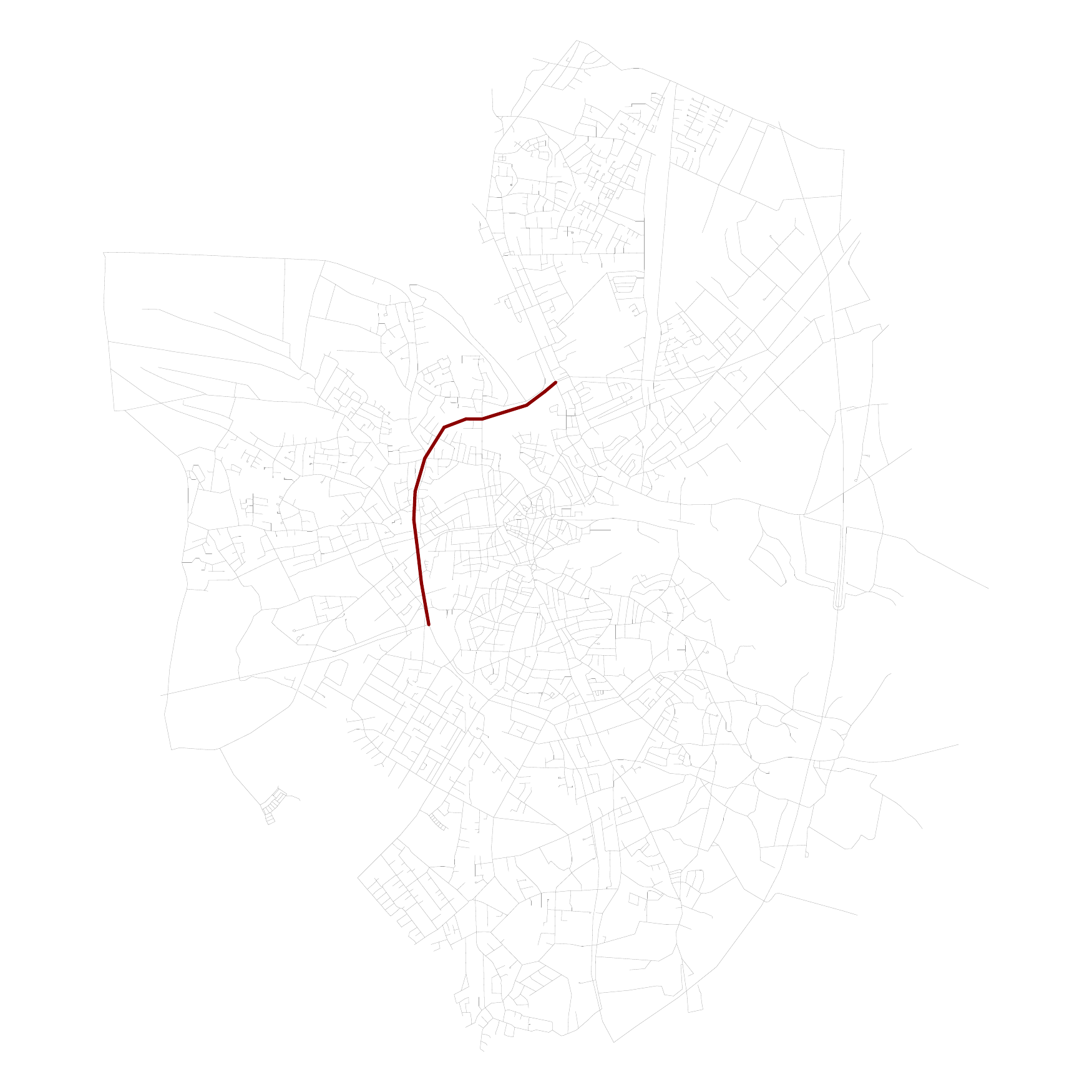}
	}
	\subfigure[Trajectory cluster 7]{
		\includegraphics[scale=0.21]{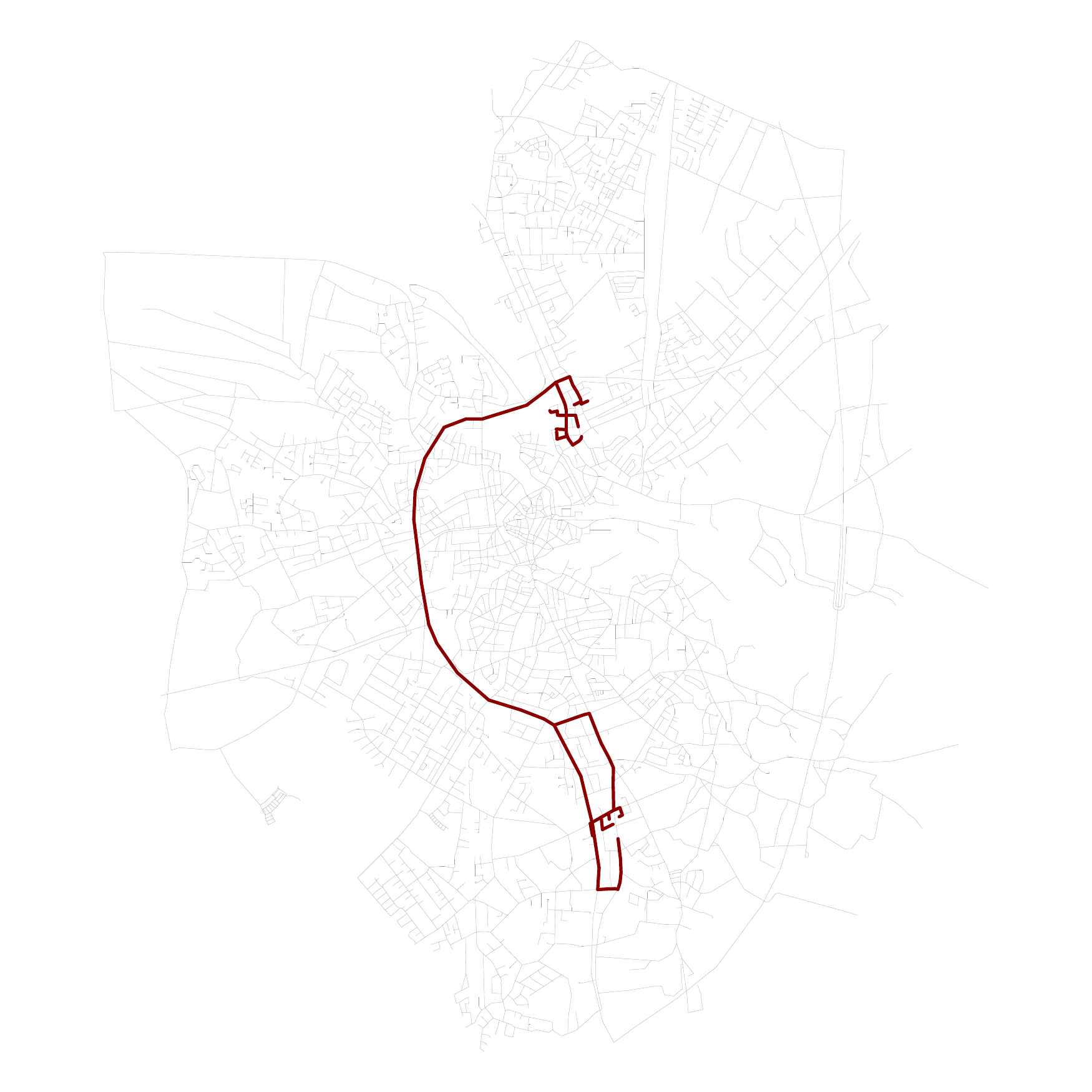}
	}
	\subfigure[Trajectory cluster 8]{
		\includegraphics[scale=0.21]{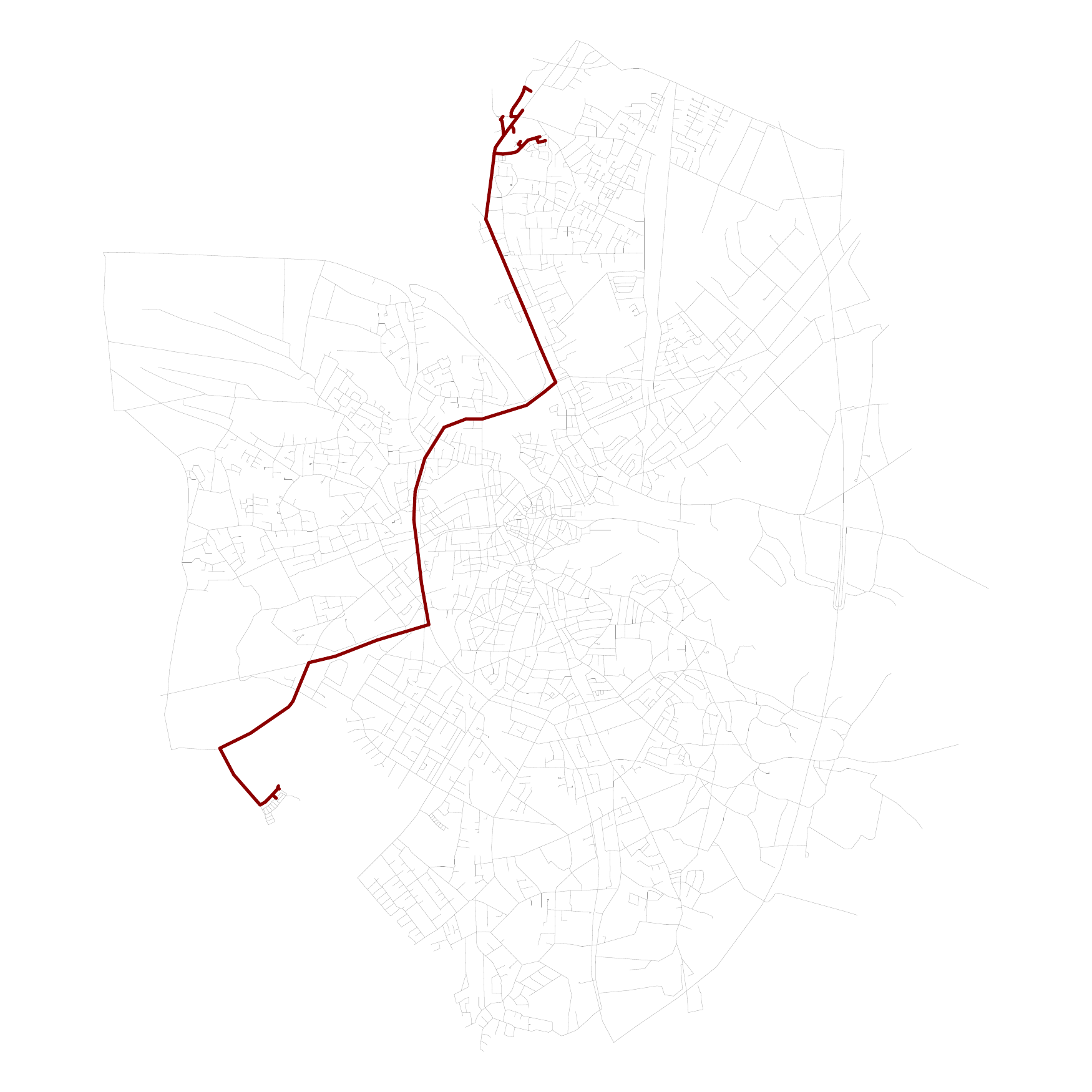}
	}
\caption{A segment cluster (a) playing the role of a hub for two different trajectory clusters ((b) and (c)) that borrow it to travel to two separate destinations.}
\label{fig:crossed}
\end{figure}

The crossed matrix does indeed reveal some interesting patterns and interactions. For instance, the fourth segments clusters is explored exclusively by two trajectory clusters. Visualizing both this segment cluster and its visiting trajectory clusters (Fig. \ref{fig:crossed}) shows that the segment cluster plays the role of a hub for these two groups of trajectories that converge to it from two different areas in order to travel to two different destinations.

Crossing trajectory clusters and segment clusters is flexible and can be done at various levels of the hierarchies of both cluster types. However, it is totally up to the user to decide the relevance of the crossed clusters. The case of the eleventh segment cluster (cf. Fig. \ref{fig:crossedmatrix}) illustrates this point: this segment cluster is very interesting since it interacts with six trajectory clusters. However, it is evident that the segment cluster contains a lot of "noise" segments which is expressed by the considerable amount of white space in the six first rows (representing the trajectory clusters) in the column representing the cluster in the crossed matrix. Consequently, drawn conclusions about the interactions between the clusters won't be very reliable. A wiser alternative would be to study the interactions between the, more refined, subclusters of this segment cluster with the six trajectory cluster it interacts with.

\subsection{Algorithmic Complexity}
\label{sec:complexity}

Let $n$ be the number of trajectories in $\mathcal{T}$ and $m$ the number of  road segments in $\mathcal{S}$. Road segments can be represented as a matrix $M$ containing $m$ rows (each representing a road segment) and $n$ columns (each corresponding to a trajectory). $m_{i,j}$ corresponds to the weight of the trajectory represented by column $j$ while inspecting the segment represented by row $i$. Using this vector model representation, comparing two road segments can then be done in $O(n)$ time complexity. Constructing the similarity graph requires $\frac{m (m-1)}{2}$ similarity calculations. Therefore, the cost of constructing the graph is $O(n m^2)$.

The similarity graph contains $m$ vertices (representing the $m$ segments of $\mathcal{S}$) and, at most, $\frac{m (m-1)}{2}$ edges. Therefore, the theoretical (maximal) complexity of the community detection algorithm used in our clustering phase is $O(m^3)$ \cite{Fortunato_2010}. However, this complexity  is rarely observed in practice where the complexity is somewhere near $O(m^2)$.

The complexity of the approach we presented in \cite{Mahrsi_2012a} can be deduced using the same reasoning: the trajectory similarity graph is constructed in $O(m n^2)$ and is clustered in $O(n^3)$ in theory ($O(n^2)$ in practice).

\section{Experimental Results}
\label{sec:experimentalresults}

In this section, we validate the effectiveness of our approach by comparing it to two alternative graph clustering techniques. First, we describe our experimental setting, including the used datasets and the evaluated algorithms in Section \ref{sec:experimentalsetting}. Then cluster quality results are presented in Section \ref{sec:results}.

\subsection{Experimental Setting}
\label{sec:experimentalsetting}

In order to validate our choice of modularity-based clustering, we compare it to two other graph clustering techniques: i. spectral clustering; and ii. label propagation clustering. In spectral clustering \cite{Luxburg_2007}, eigenvectors are extracted from the graph's Laplacian and are used to conduct a $k$-means clustering in order to partition the graph's vertices. Label propagation, on the other hand, works by labeling the vertices with unique labels and then updating the labels by majority voting in the neighborhood of the vertex \cite{Raghavan_2007}.

We compare the performances of the three algorithms on five synthetic datasets (cf. Table \ref{tab:datasets}) produced with the Brinkhoff generator \cite{Brinkhoff_2002} using the Oldenburg road network. The latter is composed of 6105 vertices and about 14070 road segments. Each dataset contains 100 trajectories visiting a various amount of road segments.

\begin{table*}[!ht]
\begin{center}
\caption{Characteristics of the five synthetic datasets.}
\label{tab:datasets}
\begin{tabular}{|c|c|c|}
\hline
 & Number of & Number of edges in\\
Dataset &  segments & the similarity graph\\
\hline
1 & 2562 & 79811\\
2 & 2394 & 100270\\
3 & 2587 & 110095\\
4 & 2477 & 87023\\
5 & 2348 & 80659\\
\hline
\end{tabular}
\end{center}
\end{table*}%

The performance of each algorithm is evaluated by measuring the quality of the segment partition $\mathcal{C}_\mathcal{S}$ it produces according to formula (\ref{eq:quality}):

\begin{equation}
	\mathcal{Q}(\mathcal{C}_\mathcal{S}) = \sum_{C \in \mathcal{C}} \frac{1}{|C|} \sum_{s_i, s_j \in C} \frac{|\{T \in \mathcal{T} : s_i \in T \wedge s_j \in T\}|}{|\{T \in \mathcal{T} : s_i \in T \vee s_j \in T\}|}
\label{eq:quality}
\end{equation}

$|C|$ is the number of segments in clusters $C$, $|\{T \in \mathcal{T} : s_i \in T \wedge s_j \in T\}|$ is the number of trajectories both road segments $s_i$ and $s_j$ while $|\{T \in \mathcal{T} : s_i \in T \vee s_j \in T\}|$ is the number of trajectories that travelled along at least one of them.

\subsection{Results}
\label{sec:results}

Contrary to the spectral clustering algorithm, the modularity-based and label propagation algorithms do not give the user the possibility to configure the number of resulting clusters: the label propagation algorithm produces just one flat partition while the modularity-based algorithm produces a partial hierarchy (i.e. a hierarchy that does not retain all the merging operations).

First, we compare modularity-based clustering and spectral clustering based on the former's optimal number of clusters (i.e. the number of clusters at the hierarchy's first level). The results are depicted in Table \ref{tab:results1}.

\begin{table}[!ht]
\begin{center}
\caption{Comparison between spectral clustering and modularity-based clustering.}
\label{tab:results1}
\begin{tabular}{|c|c|c|c|}
\hline
 & \multicolumn{2}{c|}{Clustering quality (clusters)}\\
\cline{2-3}
Dataset  & Spectral & Modularity \\
\hline
1   & 306.33 (23) & \textbf{657.20} (23)\\
2   & 254.97 (21) & \textbf{524.46} (21)\\
3 & 245.64(20) & \textbf{561.08} (20)\\
4 &249.89 (22) & \textbf{594.75} (22)\\
5 &284.74 (26) & \textbf{666.23} (26)\\
\hline
\end{tabular}
\end{center}
\end{table}

In order to compare the three algorithms at once (cf. Table \ref{tab:results2}), we proceed as follows. Since label propagation clustering produces only on partition, we configure the spectral clustering to produce the same number of clusters as this partition. As for modularity-based clustering, we choose the hierarchical level that produces the closest number of clusters to those discovered by label propagation.

\begin{table}[!ht]
\begin{center}
\caption{Cluster qualities achieved by the three algorithms on the five datasets.}
\label{tab:results2}
\begin{tabular}{|c|c|c|c|c|}
\hline
 & \multicolumn{3}{c|}{Clustering quality (clusters)}\\
\cline{2-4}
Dataset  & Label prop. & Spectral & Modularity \\
\hline
1  & 684.19 (68) & 678.81 (68) & \textbf{1614.40} (67)\\
2 & 550.66 (59) & 549.70 (59) & \textbf{1276.63} (57)\\
3  &606.45 (66) & 567.57 (66) & \textbf{1516.45} (61)\\
4 &634.63 (68) & 637.62 (68) &\textbf{1406.38} (57)\\
5 &604.97 (64) & 539.27 (64) & \textbf{1418.67} (65)\\
\hline
\end{tabular}
\end{center}
\end{table}%

From both Table \ref{tab:results1} and Table \ref{tab:results2} we can verify that, as expected, the clustering quality increases as the number of clusters increases. Results also show the superiority of modularity-based clustering over label propagation and spectral clustering and suggest that applying the former results in better and more compact clusters of road segments.

We also notice that label propagation results in a large number of clusters. This supports the observation we made in Section \ref{sec:exploration} where we claimed that flat clustering is not suitable for exploring large datasets.

\section{Related Work}
\label{sec:relatedwork}

Approaches to trajectory clustering are mainly adaptations of existing algorithms to the case of trajectories. Existing problem formulations and propositions include flock patterns \cite{Benkert_2006}, convoy patterns \cite{Jeung_2008a}, the TRACLUS partition-and-group framework \cite{Lee_2007} and the T-OPTICS and TF-OPTICS algorithms \cite{Nanni_2006}. The aforementioned algorithms use euclidean-based similarities and distances and can, therefore, be used only in the case of unconstrained trajectories. Furthermore, the majority of these approaches use density-based algorithms which suffer from two major drawbacks: i. their results are very sensitive to the parameter values; and ii. they assume that trajectories in the same cluster have a rather homogeneous density, which is rarely the case (as discussed in \cite{Roh_2010}).

Roh et Hwang \cite{Roh_2010} present a network-aware approach to clustering trajectories where the distance between trajectories in the road network is measured using shortest path calculations. A baseline algorithm, using agglomerative hierarchical clustering, as well as a more efficient algorithm, called NNCluster, are presented for the purpose of regrouping the network constrained trajectories. In \cite{Kharrat_2008}, the authors describe an approach to discovering "dense paths" or sequences of frequently traveled segments in a road network. This approach resembles our segment-based clustering although they diverge on many key aspects. For instance, the approach in \cite{Kharrat_2008} produces flat clusters using a density-based approach (which requires fine tuning) whereas ours produces a hierarchy of nested clusters and does not require parametrization. In \cite{Mahrsi_2012a}, we presented our graph-based framework to clustering network-constrained trajectories. The work described in the present paper build upon this framework as it extends it to the case of road segment clustering.

A wide variety of graph clustering algorithms was proposed in the literature, including spectral clustering \cite{Luxburg_2007}, clustering using label propagation \cite{Raghavan_2007}, etc. (complete surveys on graph clustering can be found in \cite{Schaeffer_2007,Fortunato_2010}). Among these propositions, modularity-based community detection algorithms stand out for the good results they yield in practice. We use the hierarchical modularity-based clustering implementation described in \cite{Rossi_2011} (which follows the recommendations in \cite{Noack_2009}) in our clustering step of our framework in order to detect the presence of clusters among road segments.

\section{Conclusion}
\label{sec:conclusion}
In this paper, we presented a framework for clustering road segments based on the moving object trajectories that travelled along them. The main novelty of the framework is the use of a graph representation to structure the similarity relationships and interactions between road segments.  This framework presents many advantages: i. it does not require parameters, contrary to the majority of existing approaches that are very sensitive to their threshold values; and ii. it also produces a hierarchy of nested clusters promoting exploration at various levels of granularity and detail in situations where a flat clustering approach would have produced a unique level containing a very large number of clusters. Moreover, we showed how segment clusters can be used in conjunction with the trajectory clusters we defined in \cite{Mahrsi_2012a} in order to better understand flow dynamics in the road network.

The framework, however, is not flawless. The community detection algorithm used in the clustering step can be sensitive in presence of noise (i.e. marginal road segments that do not forcefully belong to any cluster) which can degrade the quality of the discovered clusters. Also, the computational cost of the approach and the fact that it requires predisposing of all the data beforehand prohibits it from being used in a streaming context.

In future work, we will focus on alternative graph representations for trajectory data. Mainly, the use of a bipartite graph to  represent interactions between trajectories and segments. Such graphs can be partitioned using bi-clustering algorithms in order to simultaneously discover clusters of trajectories and road segments (this is done separately in the present framework) which has the main advantage of automatically crossing both types of clusters based on how they interact, thus relieving the user from this delicate task.

\bibliographystyle{splncs}
\bibliography{bibliography}

\end{document}